%% file: main.tex
\documentclass[10pt,twocolumn,letterpaper]{article}

\usepackage{xcolor}

\usepackage{graphicx}
\newcommand{\rot}[1]{\rotatebox{90}{#1}}
\usepackage{amssymb}
\usepackage{makecell}
\usepackage{multirow}
\usepackage{cuted}
\usepackage{caption}
\usepackage{array}

\usepackage[pagenumbers]{cvpr} 

\input{preamble}
\definecolor{cvprblue}{rgb}{0.21,0.49,0.74}
\usepackage[pagebackref,breaklinks,colorlinks,allcolors=cvprblue]{hyperref}

\def\paperID{132} 
\def\confName{3DV\xspace}
\def\confYear{2027\xspace}

\title{LetOccVote: Learning Weakly Supervised 3D Occupancy through Consensus}

\author{Chi Zhang{*}\\
CUHKSZ\\
{\tt\small chizhang1@link.cuhk.edu.cn}
\and
Qi Song{*}\\
Tsinghua University\\
{\tt\small songqi@mail.tsinghua.edu.cn}
\and
Feifei Li\\
CUHKSZ\\
{\tt\small feifeili1@link.cuhk.edu.cn}
\and
Jie Li {$^\dagger$}\\
Shenzhen Polytechnic University\\
{\tt\small jieli1@szpu.edu.cn}
\and
Rui Huang {$^\dagger$}\\
CUHKSZ\\
{\tt\small ruihuang@cuhk.edu.cn}
}
\begin{document}
\maketitle

\begingroup
\renewcommand{\thefootnote}{*}
\footnotetext{Equal contribution.}
\endgroup

\begingroup
\renewcommand{\thefootnote}{$^\dagger$}
\footnotetext{Corresponding authors.}
\endgroup

\begin{strip}
    \centering
    \includegraphics[width=\textwidth]{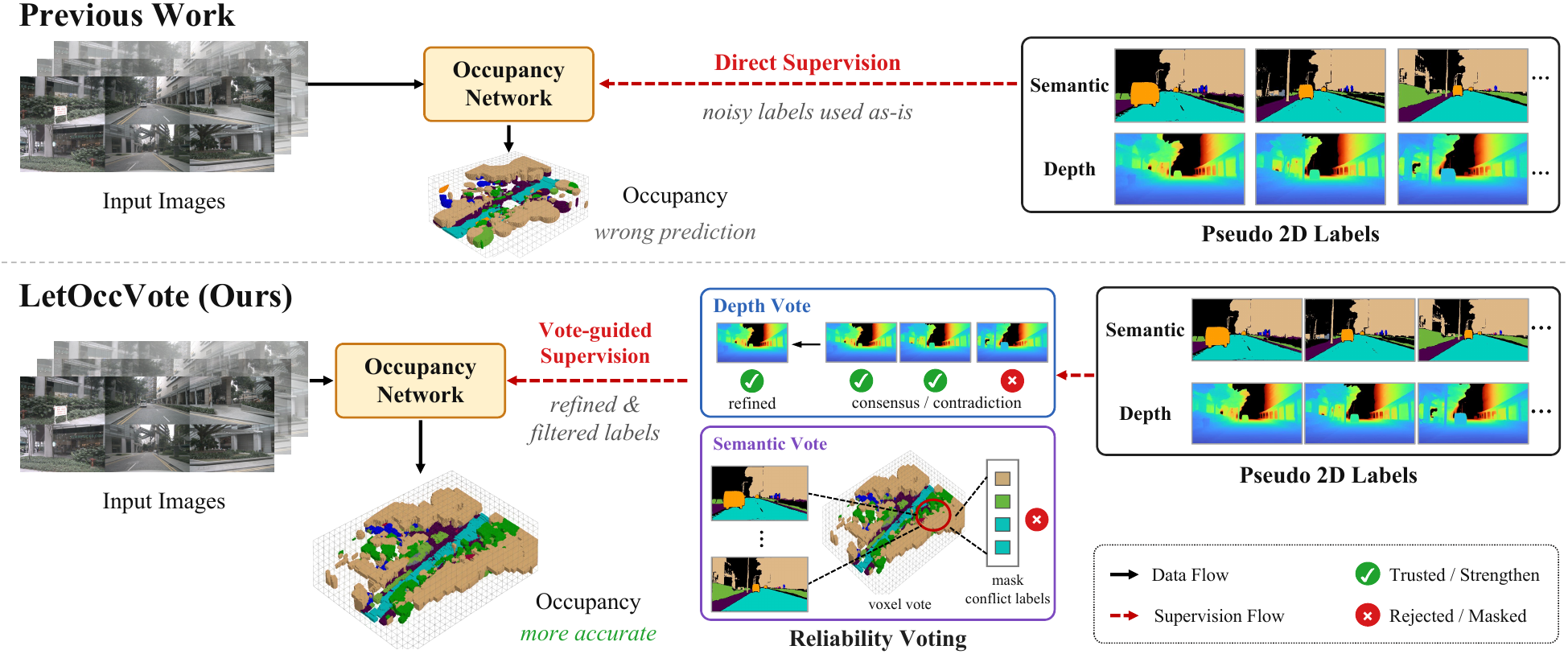}
    \captionof{figure}{\textbf{Motivation of LetOccVote.} Existing weakly supervised occupancy methods directly learn from noisy 2D pseudo-labels. LetOccVote instead exploits multi-view observation voting over depth and semantics to identify reliable and conflicting pseudo-labels, strengthening consistent supervision while masking unreliable labels for more accurate occupancy prediction.}

    \label{fig:teaser}
\end{strip}

\input{sec/0_abstract}    
\input{sec/1_intro}
\input{sec/2_relatedworks}

\input{sec/3_method}

\input{sec/4_experiments}

\input{sec/5_conclusion}

{
    \small
    \bibliographystyle{ieeenat_fullname}
    \bibliography{main}
}


\end{document}

%% file: sec/0_abstract.tex
\begin{abstract}
Weakly supervised 3D occupancy prediction reduces the reliance on costly 3D annotations by learning from 2D pseudo-labels generated by vision foundation models. However, existing methods typically use these imperfect pseudo-labels directly as supervision, making occupancy learning vulnerable to erroneous geometric and semantic targets. We observe that agreement across repeated observations provides an inexpensive and reliable cue for assessing pseudo-label reliability. Based on this observation, we propose \textbf{LetOccVote}, a weakly supervised Gaussian-based occupancy framework that leverages cross-frame voting to improve both geometric and semantic supervision. For geometry, Depth Vote exploits cross-frame geometric agreement to refine supported pseudo depth and reject contradictory estimates before volumetric lifting and depth supervision. For semantics, Semantic Vote aggregates pseudo-semantic observations in a shared 3D space to identify reliable and contested evidence, strengthening reliable semantic supervision while filtering unreliable pseudo-label segments. The entire framework is trained solely with 2D pseudo-label supervision without requiring 3D occupancy annotations. On Occ3D-nuScenes, LetOccVote achieves 53.27 IoU and 20.39 mIoU, establishing state-of-the-art performance among methods with 2D pseudo-label supervision.
\end{abstract}

%% file: sec/1_intro.tex
\section{Introduction}
\label{sec:intro}

3D occupancy captures scene geometry and semantics in a unified spatial representation, supporting downstream autonomous driving tasks such as motion planning and decision-making \cite{zheng2024occworld, yang2025driveoccWorld}. However, obtaining dense voxel-level annotations is costly and labor-intensive~\cite{wang2023openoccupancy, zhang2026visionocc_review}. Recent methods alleviate this burden by learning occupancy from 2D depth and semantic supervision through differentiable rendering \cite{pan2024renderocc, sun2024gsrender, chambon2025gaussrender}. More recently, foundation models have been used to automatically generate such supervision, further reducing the dependence on manual annotations \cite{huang2024selfocc, gan2025gaussianocc, jiang2025gausstr, boeder2025gaussianflowocc}.

Despite this progress, pseudo supervision remains inherently unreliable. Pseudo depth can exhibit geometric errors and cross-frame inconsistencies, while pseudo semantics may confuse visually similar categories. Existing methods have already exploited complementary information across different sources or observations to improve weakly supervised occupancy learning. For example, GaussTR leverages knowledge from multiple foundation models to align and enrich 3D representations \cite{jiang2025gausstr}, while GaussianFlowOcc exploits temporal observations through both attention mechanism and cross-frame consistency \cite{boeder2025gaussianflowocc}. However, in these methods, agreement across different sources or observations is mainly used to improve representation learning or impose consistency constraints, rather than to explicitly evaluate the reliability of pseudo-labels. As a result, unreliable pseudo-labels can still act as direct optimization targets and interfere with the learning of accurate geometry and semantics. Explicitly distinguishing reliable supervision from unreliable pseudo-labels therefore remains an important yet underexplored problem.

Our key observation is that agreement across repeated observations can serve not only as additional information, but also as direct evidence of pseudo-label reliability. The same physical location should exhibit consistent geometry and semantics when observed from different viewpoints and timestamps, while contradictory observations may indicate potentially unreliable supervision. Based on this observation, we propose \textbf{LetOccVote}, a weakly supervised Gaussian-based occupancy framework. It leverages agreement across multiple observations to identify reliable pseudo supervision, as illustrated in Fig.~\ref{fig:teaser}. For geometry, Depth Vote exploits cross-frame geometric agreement to refine supported pseudo depth and reject contradictory estimates before volumetric lifting and depth supervision. For semantics, Semantic Vote aggregates pseudo-semantic observations in a shared 3D space to identify reliable and contested evidence, strengthening reliable semantic supervision while filtering unreliable pseudo-label segments. By using multi-observation consensus, LetOccVote reduces the adverse influence of noisy pseudo-labels and enables more robust occupancy learning under imperfect supervision.

\begin{itemize} 
\item We develop a weakly supervised Gaussian-based occupancy framework that integrates Depth Vote with volumetric lifting, enabling image features from repeated observations to be aggregated into a unified 3D representation for Gaussian-based occupancy prediction.
\item We further apply Depth Vote and introduce Semantic Vote to improve the reliability of pseudo supervision. Depth Vote refines or rejects geometrically inconsistent pseudo depth, while Semantic Vote reinforces reliable semantic evidence and masks unreliable pseudo-label segments.
\item Extensive experiments on Occ3D-nuScenes demonstrate that LetOccVote achieves state-of-the-art performance among methods trained with 2D pseudo-label supervision. Ablation studies further validate the effectiveness and robustness of the proposed depth and semantic voting strategies.
\end{itemize}

%% file: sec/2_relatedworks.tex
\section{Related Work}
\label{sec:related}

\subsection{Occupancy Supervised  by 2D Pseudo-Labels}

Camera-based 3D occupancy prediction has been widely studied with volumetric, sparse, and Gaussian-based representations \cite{cao2022monoscene, wei2023surroundocc, huang2023tpvformer, tang2024sparseocc, huang2024gaussianformer, huang2025gaussianformerv2}. However, these methods rely on 3D supervision that is costly to annotate and acquire. To alleviate this burden, recent studies have explored learning 3D occupancy from supervision available in the image space. RenderOcc \cite{pan2024renderocc} supervises rendered occupancy predictions with 2D depth and semantic labels, while subsequent rendering-based approaches exploit image-space, geometric, and temporal cues to further reduce the dependence on 3D annotations \cite{zhang2023occnerf, huang2024selfocc, boeder2025occflownet,liu2025letoccflow}. GaussianOcc \cite{gan2025gaussianocc} further introduces 3D Gaussians into this rendering-based framework, providing a more efficient intermediate representation for occupancy learning.

More recently, pretrained foundation models have further reduced annotation requirements by providing scalable supervision without task-specific manual labeling. LangOcc \cite{boeder2025langocc} and GaussTR \cite{jiang2025gausstr} transfer pretrained visual knowledge into 3D representations through rendering-based alignment, whereas GaussianFlowOcc \cite{boeder2025gaussianflowocc} directly employs foundation models to generate pseudo depth and semantic labels for weakly supervised occupancy learning. EasyOcc \cite{hayes2025easyocc} and ShelfOcc \cite{boeder2026shelfocc} consolidate foundation-model predictions into 3D pseudo-labels for direct occupancy supervision, but require additional preprocessing and cross-frame aggregation. We instead retain image-space pseudo-labels, which are easier to generate at scale and preserve observation-level evidence for assessing supervision reliability across repeated observations.



\subsection{Volumetric Representation in Occupancy Prediction}

Volumetric representations are widely used in camera-based occupancy prediction to organize multi-view image features before 3D decoding \cite{wei2023surroundocc,huang2023tpvformer,tang2024sparseocc}. More recent Gaussian-based methods adopt a more direct image-to-Gaussian pipeline. The GaussianFormer series \cite{huang2024gaussianformer,huang2025gaussianformerv2}, GaussTR \cite{jiang2025gausstr}, and GaussianFlowOcc \cite{boeder2025gaussianflowocc} initialize Gaussian queries and update them by sampling image-view features.

Different from these approaches, we retain an intermediate volumetric representation before Gaussian decoding. Image features are lifted into a shared 3D space using pseudo depth, allowing repeated observations to be spatially aligned before Gaussian sampling. This makes geometric correspondence directly available during feature construction, but also makes the lifted volume sensitive to pseudo-depth errors. We therefore use geometric agreement across repeated observations to refine or reject unreliable depth estimates before volumetric lifting.

\subsection{Robust Learning from Pseudo-Labels} 

Pseudo-labels inevitably contain errors that may introduce misleading supervision. A common strategy is to estimate prediction quality and selectively retain reliable pseudo-labels, as explored in 3D object detection by 3DIoUMatch, ST3D, and HSSDA \cite{wang2021_3dioumatch, yang2021st3d, liu2023hssda}. Beyond filtering, U2PL \cite{wang2022u2pl} separates reliable and unreliable pixels according to prediction entropy and exploits the latter as negative samples, while ELN \cite{kwon2022eln} learns an auxiliary error localization network to identify and suppress erroneous pseudo-label regions. Other studies further show that pseudo-label errors can exhibit structured patterns. DebiasMatch \cite{Wang2022debiasmatch} mitigates class bias induced by imbalanced pseudo-label distributions, while DPL \cite{zhang2024dpl} addresses reliability disparities between 2D and 3D pseudo-label attributes and the optimization conflicts caused by noisy depth supervision.

These studies demonstrate that robust pseudo-label learning can benefit from both reliability-aware selection and explicit treatment of unreliable supervision. Building on this insight, we use agreement across repeated observations to assess pseudo depth and semantic supervision according to their distinct reliability patterns.

%% file: sec/3_method.tex
 \section{Method}
\label{sec:method}

\begin{figure*}[t]
  \centering
   \includegraphics[width=1.0\linewidth]{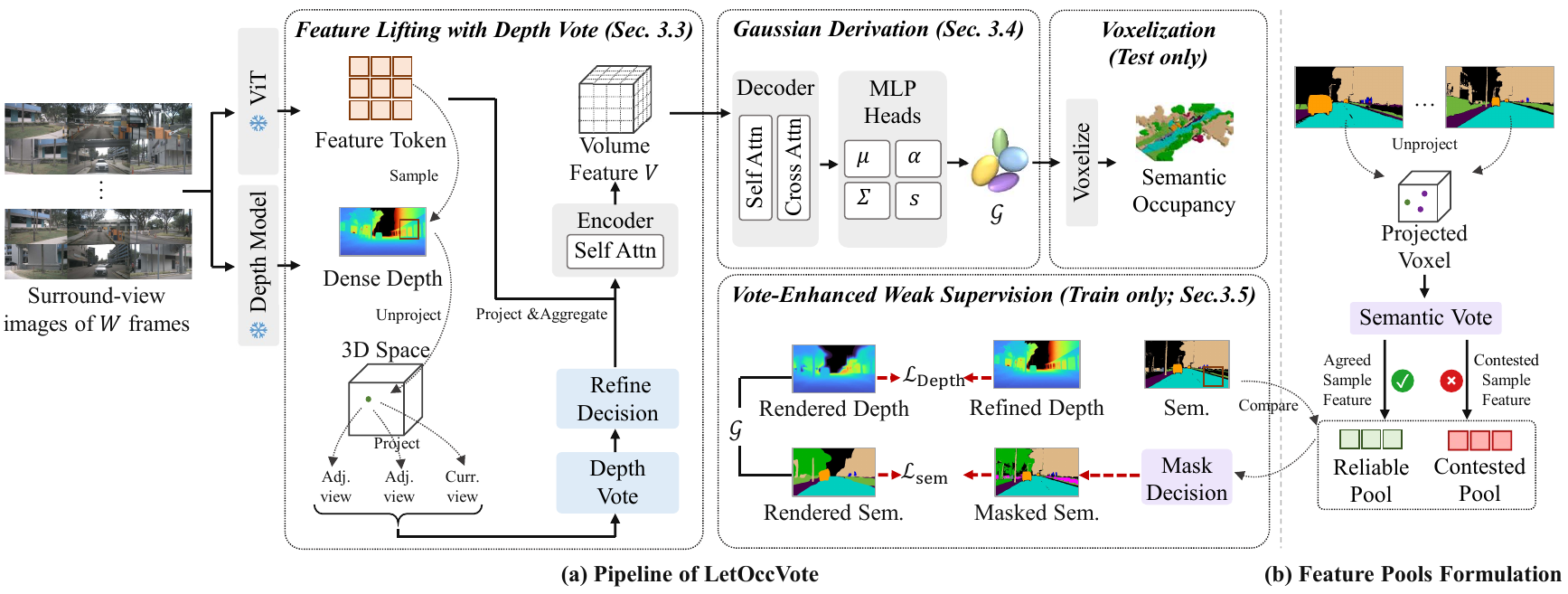}

   \caption{\textbf{Overview of the LetOccVote.} \textbf{Fig. (a)} The overall framework lifts multi-frame observations into a 3D volume with Depth Vote, derives 3D Gaussians, and applies vote-enhanced weak supervision during training. \textbf{Fig. (b)} Semantic Vote separates agreed and contested samples to construct feature pools, which guide the masking of unreliable semantic segments in (a).}
   \label{fig:pipeline}
\end{figure*}

\subsection{Problem Formulation}
Let $\mathcal{T}$ denote a temporal sequence of $W$ surround-view keyframes associated with the current frame $t=0$, and let $I_{t,n}\in\mathbb{R}^{3\times H_I\times W_I}$ denote the image from camera $n$ at frame $t\in\mathcal{T}$, with known camera parameters and ego poses. Our goal is to predict the semantic occupancy $O\in\mathcal{C}^{X\times Y\times Z}$ of the current frame on a fixed ego-centric voxel grid. We represent the scene with a set of semantic 3D Gaussians $\mathcal{G}=\left\{\left(\boldsymbol{\mu}_i,\boldsymbol{\Sigma}_i,\alpha_i,\mathbf{s}_i\right)\right\}_{i=1}^{N_G}$, where $\boldsymbol{\mu}_i$, $\boldsymbol{\Sigma}_i$, $\alpha_i$, and $\mathbf{s}i$ denote the center, covariance, opacity, and semantic prediction of the $i$-th Gaussian, respectively. During training, no 3D occupancy annotations are available. Instead, each image is associated with a semantic pseudo-label map $\tilde{S}_{t,n}$ and a metric depth pseudo-label map $\tilde{D}_{t,n}$ generated by off-the-shelf 2D foundation models, which provide weak supervision for learning $\mathcal{G}$. At inference, the Gaussians are voxelized to obtain the semantic occupancy $O$.

\subsection{Overview}

The pipeline of LetOccVote is illustrated in Fig.~\ref{fig:pipeline} (a). We first extract image features $F$. Before feature lifting, Depth Vote exploits geometric agreement across observations to refine consistent pseudo-depth estimates and suppress inconsistent ones. Using the refined depth and camera geometry, the image features are projected and aggregated into a shared 3D volume, which is further refined by a 3D encoder to obtain the volume feature $V \in \mathbb{R}^{C\times X\times Y\times Z}$. A 3D decoder then queries $V$ to produce the semantic Gaussian representation $\mathcal{G}$. During training, Semantic Vote is applied to the semantic pseudo-labels before semantic supervision to identify reliable and contested regions, thereby strengthening reliable supervision while suppressing unreliable labels. The temporal sequence is centered at the current frame and includes both past and future frames during training for richer supervision. At inference, we use the same number of frames but only the current and preceding frames to ensure causal prediction.

\subsection{Feature Lifting with Depth Vote}

Image-space supervision mainly constrains directly observed surfaces, whereas occupancy prediction requires a unified 3D scene representation. We therefore lift temporal multi-view features into a shared 3D volume. Since pseudo depth directly determines where image features are placed in 3D, local depth errors can introduce misplaced features and corrupt the resulting volume. To assess pseudo-depth reliability without 3D annotations, we exploit geometric agreement across repeated observations: a valid surface point should correspond to a consistent 3D position across views. Since a single cross-frame discrepancy may result from local depth noise, we aggregate multiple depth comparisons and make a correction only when at least $k$ observations provide consistent evidence.

For each depth sample associated with an image feature, we treat its pseudo depth as a candidate surface hypothesis and verify it against the remaining frames in $\mathcal{T}$ before lifting. Specifically, the sample is first unprojected into the current ego 3D space and then reprojected into each comparison frame, where the pseudo depth at the corresponding image location is queried. To reduce image-wise scale bias, the comparison depth is first aligned to the target observation using a robust median depth ratio estimated over valid projected points. Let $z$ denote the reprojected depth and $\bar{d}$ the aligned pseudo depth in a comparison view. Their geometric consistency is categorized according to
\begin{equation}
\begin{aligned}
\text{Agreement:}\quad
& |z-\bar{d}| \leq \tau(z),\\
\text{Contradiction:}\quad
& z < \bar{d}-\tau(z),\\
\text{Abstention:}\quad
& \text{otherwise}.
\end{aligned}
\label{eq:depth_evidence}
\end{equation}
Here, $\tau(z)$ is a depth-adaptive tolerance that increases with distance. Agreement indicates that the comparison supports the candidate surface location, whereas contradiction indicates that this location lies in front of the observed surface. If the candidate location lies far behind the observed surface, it may be occluded, and we therefore treat this case as an abstention. Out-of-view projections and invalid depth measurements are likewise treated as abstentions. Camera-level evidence within the same temporal frame is merged into a single frame-level vote, such that each frame contributes at most one vote.

We then consolidate the frame-level evidence using a $k$-frame voting rule over the $W$-frame temporal window to determine the depth used for feature lifting. Let $N_{\mathrm{agr}}$ and $N_{\mathrm{con}}$ denote the numbers of agreeing and contradicting frames, respectively. The refined depth used for lifting is defined as
\begin{equation}
\hat{D}_{t,n} =
\begin{cases}
\mathcal{R}\!\left(\tilde{D}_{t,n}\right),
& N_{\mathrm{agr}} \geq k,\\
0,
& t\neq 0,\; N_{\mathrm{con}} \geq k,\; N_{\mathrm{agr}} = 0,\\
\tilde{D}_{t,n},
& \text{otherwise},
\end{cases}
\label{eq:depth_vote}
\end{equation}
where $\mathcal{R}(\cdot)$ denotes the consensus-based depth correction derived from the agreeing frames together with the original depth estimate. A depth sample receiving agreement from at least $k$ frames is refined and lifted at the corrected 3D location. In contrast, a sample contradicted by at least $k$ frames without any agreement is invalidated and excluded from lifting, while insufficient or ambiguous evidence leaves the original pseudo depth unchanged. For the current frame, rejection is disabled, although its depth can still be refined when sufficient agreement is available. During training, the refined depth is also used to supervise the rendered depth, as detailed in Sec.~\ref{sec:depthsup}.

Using the refined depth $\hat{D}_{t,n}$, features from all cameras and temporal frames are unprojected into the current ego-centric coordinate system and splatted into a shared voxel grid. Features assigned to the same voxel are averaged to form an intermediate volumetric representation. A 3D encoder with deformable self-attention further refines the voxel features to incorporate spatial context. The encoded features form the final volume feature $V\in\mathbb{R}^{C\times X\times Y\times Z}$, which is subsequently queried by the Gaussian decoder.

\subsection{Gaussian Derivation}

Given the refined volume feature $V$, we sample a fixed number of query reference positions from occupied voxels and use their voxel centers as the initial 3D references. The queries interact with $V$ through the decoder, which progressively refines their reference positions and query features.

The refined references determine the Gaussian centers $\boldsymbol{\mu}_i$, while lightweight MLP heads predict the remaining Gaussian attributes, including the opacity $\alpha_i$, covariance $\boldsymbol{\Sigma}_i$, and semantic logits $\mathbf{s}_i$, yielding the semantic Gaussian set $\mathcal{G}$.



\subsection{Vote-Enhanced Weak Supervision}

\subsubsection{Refined Depth Supervision}
\label{sec:depthsup}
The consensus-refined depth $\hat{D}_{t,n}$ in Eq.~\ref{eq:depth_vote} is also used as the weak geometric supervision for Gaussian rendering. Specifically, the predicted Gaussian representation $\mathcal{G}$ is rendered into the current and neighboring frames, and the rendered depth $D^{\mathrm{rend}}_{t,n}$ is supervised by the corresponding refined depth. The depth objective is defined as
\begin{equation}
\begin{aligned}
\mathcal{L}_{\mathrm{depth}}
=
\sum_{(t,n)} w_t
\Big[
&\mathcal{L}_{\mathrm{SiLog}}
\left(D^{\mathrm{rend}}_{t,n},\hat{D}_{t,n}\right) \\
&+
\lambda_{\mathrm{L1}}
\mathcal{L}_{1}
\left(D^{\mathrm{rend}}_{t,n},\hat{D}_{t,n}\right)
\Big].
\end{aligned}
\label{eq:depth_supervision}
\end{equation}
The losses are evaluated only on valid depth targets, and $w_t$ controls the contribution of frame $t$ to the depth supervision. Samples invalidated by Eq.~\ref{eq:depth_vote} are excluded from supervision. Thus, the same refined depth consistently guides both feature lifting and rendering-based optimization.

\subsubsection{Semantic Supervision}

Unlike depth pseudo-labels, semantic pseudo-labels exhibit class-dependent and recurring category confusions that cannot be addressed solely through geometric consistency. We therefore exploit semantic agreement across observations to distinguish reliable and potentially unreliable supervision rather than treating all pseudo-labels equally. To this end, Semantic Vote aggregates semantic observations in 3D to obtain consistent voxel-wise evidence and expose conflicts between pseudo-label classes and voted classes. Consistent votes provide additional supervision for Gaussian semantics, while agreed and contested observations are accumulated in the frozen feature space to characterize their respective appearance patterns. These patterns are subsequently used to identify and suppress semantic pseudo-labels that resemble contested observations. In this way, Semantic Vote strengthens reliable semantic cues while reducing the influence of potentially unreliable supervision.

\paragraph{Semantic Vote.}
Using the refined depth $\hat{D}_{t,n}$, we project the semantic pseudo-labels $\tilde{S}_{t,n}$ from all cameras and frames in $\mathcal{T}$ into the shared ego-centric voxel grid. Let $\mathcal{Y}_v$ denote the set of valid semantic candidates projected into voxel $v$. The voted class $y_v$ is determined by majority voting over $\mathcal{Y}_v$, while the agreement ratio $A_v$ measures the fraction of candidates supporting this decision.

\paragraph{Reliable and Contested Feature Pools.}
As illustrated in Fig.~\ref{fig:pipeline} (b), we use the voxel voting results to separate frozen backbone features into reliable and contested pools. Each feature token is associated with a voxel using the refined depth. For a token with pseudo-label $c$, if its pseudo-label agrees with the voted class $y_v$ and the agreement ratio $A_v$ exceeds a reliability threshold, it is added to the Reliable Pool $F^{\mathrm{rel}}_c$. If its pseudo-label disagrees with $y_v=j$, the token is instead added to the corresponding Contested Pool $F^{\mathrm{ctd}}_{c,j}$. Such disagreement indicates a semantic conflict across observations without determining which of the two classes is correct.

We summarize the Reliable Pool of class $c$ by a reference feature $\mathbf{r}_c$, while its Contested Pools are represented by a set of reference features $\mathbf{Q}_c^{\neg c}$. Specifically,
\begin{equation}
\mathbf{r}_c =
\operatorname{Norm}\!\left(
\sum_{\mathbf{f}\in F^{\mathrm{rel}}_c} \mathbf{f}
\right),
\mathbf{Q}_c^{\neg c}
=
\left\{\operatorname{Norm}\!\left(\sum_{\mathbf{f}\in F^{\mathrm{ctd}}_{c,j}} \mathbf{f}\right)\right\}_{j\neq c}.
\label{eq:semantic_feature_refs}
\end{equation}
Here, $\mathbf{r}_c$ characterizes the appearance pattern supported by consistent semantic votes, whereas each element of $\mathbf{Q}_c^{\neg c}$ characterizes a contested appearance pattern associated with a different voted class. The feature statistics of these pools are accumulated during an initial training stage and subsequently frozen for the semantic masking described next.

\vspace{-4pt}

\paragraph{Semantic Masking and Auxiliary Voted Supervision.}

We use the frozen reference features to identify pseudo-labeled regions that resemble previously observed semantic conflicts. For a token feature $\mathbf{f}$ with pseudo-label class $c$, we compare its similarity to the reliable reference feature $\mathbf{r}_c$ with its maximum similarity to the contested references $\mathbf{Q}_c^{\neg c}$:
\begin{equation}
\Delta(\mathbf{f},c)
=
\max_{\mathbf{q}\in\mathbf{Q}_c^{\neg c}}
\operatorname{sim}(\mathbf{f},\mathbf{q})
-
\operatorname{sim}(\mathbf{f},\mathbf{r}_c),
\label{eq:semantic_conflict_score}
\end{equation}
where $\operatorname{sim}(\cdot,\cdot)$ denotes cosine similarity. A token is considered contested when $\Delta(\mathbf{f},c)$ exceeds a predefined margin. The token-level decisions are aggregated within each connected segment of the same pseudo-label class, and a segment is masked when a sufficient fraction of its scored tokens are contested. Masked pseudo-labels are excluded from the semantic loss rather than reassigned to another class, while the retained pseudo-labels continue to provide their original semantic supervision.

To prevent semantic masking from disproportionately reducing the supervision of frequently masked classes, we compensate the class weights according to the smoothed fraction of retained pseudo-labels. Specifically, for class $c$, the adjusted weight is $w'_c=w_c/\rho_c$, where $\rho_c$ denotes the smoothed fraction of class-$c$ pseudo-labels that remain unmasked. This compensation allows the masking to primarily change which samples provide supervision rather than uniformly weakening the contribution of a semantic class.

High-confidence voxel votes are additionally used as sparse auxiliary supervision. Gaussian primitives whose centers fall within voxels with sufficient voting support and high semantic agreement are directly supervised by the corresponding voted class $y_v$. This auxiliary constraint complements the rendering-based semantic supervision without replacing the original pseudo-labels. The semantic supervision is formulated as
\begin{equation}
\begin{aligned}
\mathcal{L}_{\mathrm{sem}} =
& \sum_{(t,n)} w_t\,\mathcal{L}_{\mathrm{CE}}\left(S^{\mathrm{rend}}_{t,n},\tilde{S}_{t,n};M_{t,n},w'\right) \\
& + \lambda_{\mathrm{vote}}\sum_{i\in\mathcal{G}_{\mathrm{vote}}}\mathcal{L}_{\mathrm{CE}}\left(\mathbf{s}_i,y_{v_i}\right).
\end{aligned}
\label{eq:semantic_supervision}
\end{equation}
Here, $M_{t,n}$ denotes the semantic keep mask obtained from the segment-level masking decision, and $w'$ denotes the class weights adjusted by the retention compensation. $\mathcal{G}_{\mathrm{vote}}$ denotes the Gaussian primitives whose centers fall within sufficiently supported, high-agreement voted voxels, and $y_{v_i}$ is the voted class of the voxel containing Gaussian $i$.



%% file: sec/4_experiments.tex
\section{Experiments}
\begin{figure*}[t]
  \centering
   \includegraphics[width=1.0\linewidth]{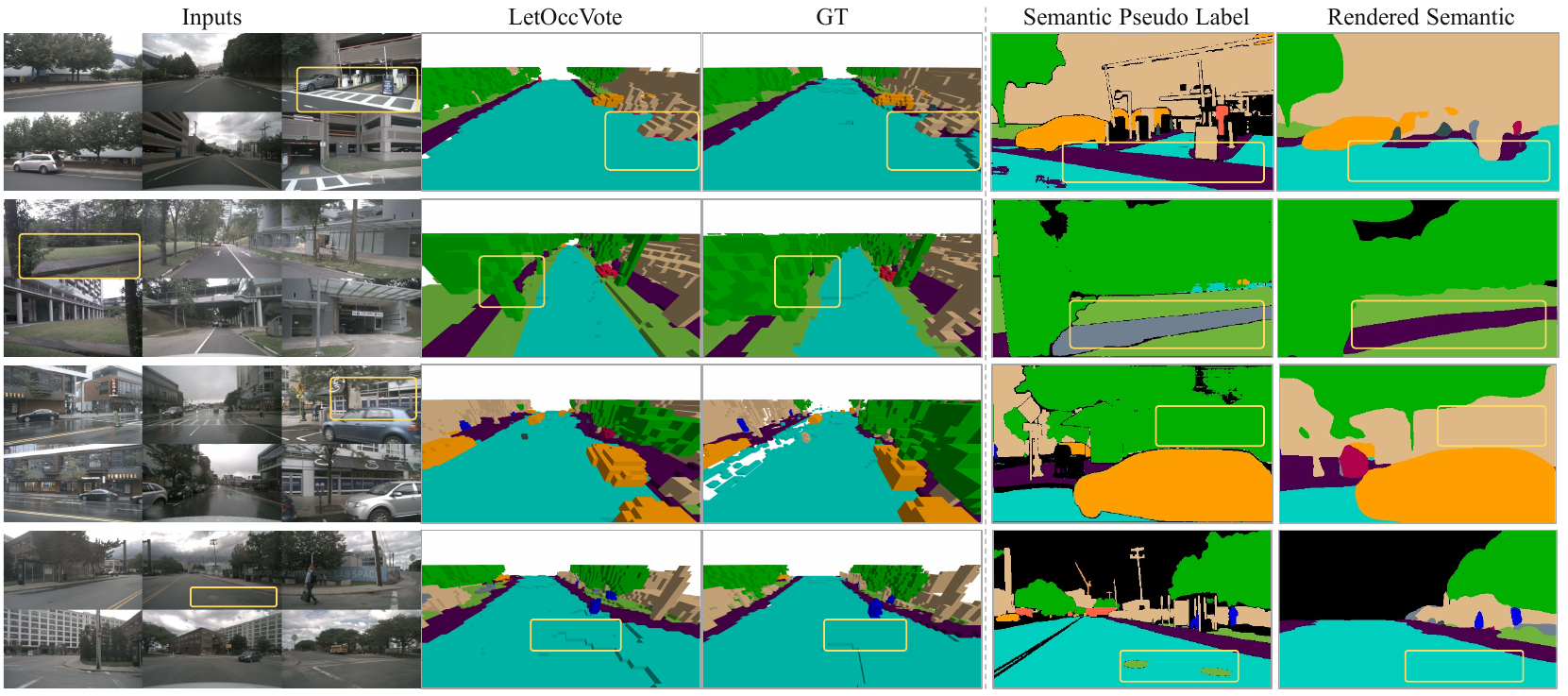}

   \caption{Qualitative results on the Occ3D-nuScenes validation set. The yellow boxes highlight the matched regions across columns.
}
   \label{fig:main}
\end{figure*}

\subsection{Dataset and Metrics} We conduct experiments on the Occ3D-nuScenes benchmark~\cite{tian2023occ3d, caesar2020nuscenes}. Our model is trained without 3D occupancy or LiDAR supervision, using only 2D pseudo labels generated by Grounded-SAM~\cite{ren2024groundedsam} for semantic segmentation and Metric3D-v2~\cite{hu2024metric3dv2} for monocular depth estimation. We evaluate the model using both IoU-based and RayIoU-based \cite{liu2024sparseocc_rayiou} metrics. For IoU-based evaluation, we report semantic mIoU and occupancy IoU. Semantic mIoU is computed over the semantic categories defined by Occ3D-nuScenes, while occupancy IoU treats all non-free classes as occupied and evaluates them against the free class. Unless otherwise specified, all categories provided by Occ3D-nuScenes are included in the mIoU computation. Since the \emph{others} and \emph{other flat} categories are not well defined, some methods exclude these two categories when computing mIoU. We mark such results with $^*$ for clarity. We additionally report RayIoU to evaluate occupancy predictions along camera rays. All IoU-based evaluations follow the official Occ3D-nuScenes protocol and are performed within the camera visibility mask. 

\subsection{Setup} 
We adopt DINOv2~\cite{oquab2023dinov2} as the image backbone and keep it frozen throughout training. Input images are resized to $504\times896$. We train the model for 10 epochs using AdamW~\cite{loshchilov2018adamw} with an initial learning rate of $1\times10^{-4}$ and a weight decay of $5\times10^{-3}$, together with a cosine annealing learning-rate schedule. We report results with two model configurations in Table~\ref{tab:main}: Base (B) and Large (L). The Base configuration uses 6,400 Gaussians with $k=2$ and a temporal window size of $W=5$, while the Large configuration uses 12,800 Gaussians with $k=4$ and $W=9$. Unless otherwise specified, all ablation studies are conducted using the Base configuration to ensure consistent and computationally efficient comparisons.

\subsection{Main Results}

We compare LetOccVote with existing weakly supervised 3D occupancy prediction methods based on 2D pseudo-label supervision in Table~\ref{tab:main} and Table~\ref{tab:rayiou}. Among the compared approaches, GaussTR~\cite{jiang2025gausstr} and GaussianFlowOcc~\cite{boeder2025gaussianflowocc} provide particularly relevant baselines, as both also use depth information during inference. As reported in Table~\ref{tab:main}, LetOccVote-B/L achieve 51.45/53.27 IoU and 22.07/23.11 mIoU$^*$, consistently outperforming previous methods. 

The improvements are broadly distributed across semantic categories: LetOccVote-L achieves the best result on 11 of the 15 classes, while the two variants together rank first on 14 classes. Particularly pronounced gains over previous methods are observed for vegetation, construction vehicle, sidewalk, car, and manmade, covering both foreground objects and large-scale scene structures.

Table~\ref{tab:rayiou} further evaluates geometric prediction quality. LetOccVote-B/L achieve 19.86/20.91 mRayIoU, and LetOccVote-L obtains the best performance at all evaluated RayIoU thresholds. These results indicate that the geometric improvements remain consistent under different matching tolerances, demonstrating more accurate and stable occupancy reconstruction.

Qualitative results are presented in Fig.~\ref{fig:main}. We compare the predicted occupancy with the ground truth in the left part of the figure, and additionally show the semantic pseudo-labels and Gaussian-rendered semantics from the corresponding views on the right, for clearer inspection. Overall, LetOccVote faithfully reconstructs both scene geometry and semantic layouts. Benefiting from dense 2D supervision, the model can also capture valid semantic occupancy in regions where the ground-truth annotation is incomplete, as illustrated in row 3. The right part further reveals diverse errors in the pseudo-labels: the first two rows contain incorrect category assignments, the third row exhibits missing labels caused by partial occlusion, and the last row shows false detections induced by appearance or texture variations. Despite these different forms of pseudo-label noise, the rendered semantics recover the corresponding scene structures more accurately, demonstrating that LetOccVote can effectively suppress unreliable supervision rather than directly inheriting its errors.

\begin{table*}[htbp]
  \centering
  \caption{Occupancy prediction results on the Occ3D-nuScenes validation set.  The best and second-best results for each semantic class are highlighted in \textbf{bold} and \underline{underline}, respectively.}
  \label{tab:main}
  \resizebox{\textwidth}{!}{
  \begin{tabular}{l|c|c|cccccccccccccccc}
    \hline
Method & IoU & mIoU &
\rot{\textcolor[rgb]{0.439,0.502,0.565}{\rule{1em}{1em}} barrier} &
\rot{\textcolor[rgb]{0.863,0.078,0.235}{\rule{1em}{1em}} bicycle} &
\rot{\textcolor[rgb]{1.000,0.498,0.314}{\rule{1em}{1em}} bus} &
\rot{\textcolor[rgb]{1.000,0.620,0.000}{\rule{1em}{1em}} car} &
\rot{\textcolor[rgb]{0.914,0.588,0.275}{\rule{1em}{1em}} cons. veh.} &
\rot{\textcolor[rgb]{1.000,0.239,0.388}{\rule{1em}{1em}} motorcycle} &
\rot{\textcolor[rgb]{0.000,0.000,0.902}{\rule{1em}{1em}} pedestrian} &
\rot{\textcolor[rgb]{0.184,0.310,0.310}{\rule{1em}{1em}} traffic cone} &
\rot{\textcolor[rgb]{1.000,0.549,0.000}{\rule{1em}{1em}} trailer} &
\rot{\textcolor[rgb]{1.000,0.384,0.275}{\rule{1em}{1em}} truck} &
\rot{\textcolor[rgb]{0.000,0.812,0.749}{\rule{1em}{1em}} drive. surf.} &
\rot{\textcolor[rgb]{0.294,0.000,0.294}{\rule{1em}{1em}} sidewalk} &
\rot{\textcolor[rgb]{0.439,0.706,0.235}{\rule{1em}{1em}} terrain} &
\rot{\textcolor[rgb]{0.871,0.722,0.529}{\rule{1em}{1em}} manmade} &
\rot{\textcolor[rgb]{0.000,0.686,0.000}{\rule{1em}{1em}} vegetation} \\
    \hline
    SelfOcc~\cite{huang2024selfocc} & 45.01 & 9.30 & 0.15 & 0.66 & 5.46 & 12.54 & 0.00 & 0.80 & 2.10 & 0.00 & 0.00 & 8.25 & 55.49 & 26.30 & 26.54 & 14.22 & 5.60 \\
    OccNeRF~\cite{zhang2023occnerf} & 22.81 & 9.53 & 0.83 & 0.82 & 5.13 & 12.49 & 3.50 & 0.23 & 3.10 & 1.84 & 0.52 & 3.90 & 52.62 & 20.81 & 24.75 & 18.45 & 13.19 \\
    DistillNeRF~\cite{wang2024distillnerf} & 29.11 & 8.93 & 1.35 & 2.08 & 10.21 & 10.09 & 2.56 & 1.98 & 5.54 & 4.62 & 1.43 & 7.90 & 43.02 & 16.86 & 15.02 & 14.06 & 15.06 \\
    GaussianOcc~\cite{gan2025gaussianocc} & - & 9.94 & 1.79 & 5.82 & 14.58 & 13.55 & 1.30 & 2.82 & 7.95 & 9.76 & 0.56 & 9.61 & 44.59 & 20.10 & 17.58 & 8.61 & 10.29 \\
    
    GaussTR~\cite{jiang2025gausstr} & 44.54 & 12.27 & 6.50 & 8.54 & 21.77 & 24.27 & 6.26 & 15.48 & 7.94 & 1.86 & \underline{6.10} & 17.16 & 36.98 & 17.21 & 7.16 & 21.18 & 9.99 \\
    GaussianFlowOcc~\cite{boeder2025gaussianflowocc} & 46.91 & 17.08* & 6.75 & 9.68 & 18.98 & 17.15 & 4.19 & 11.78 & 9.27 & \textbf{10.30} & 1.83 & 12.33 & 61.03 & 31.17 & 34.78 & 14.66 & 12.40 \\ 
    \hline
    LetOccVote-B & \underline{51.45} & \underline{19.49}/\underline{22.07}* & \underline{7.68} & \underline{12.95} & \textbf{23.46} & \underline{26.90} &
    \textbf{13.78} & \underline{15.87} & \underline{13.01} & \underline{9.90} &
    2.91 & \textbf{18.99} & \underline{63.49} & \underline{34.14} &
    \underline{37.70} & \underline{24.26} & \underline{26.06} \\

    LetOccVote-L & \textbf{53.27} & \textbf{20.39}/\textbf{23.11}* &
    \textbf{8.36} & \textbf{13.12} & \underline{22.02} & \textbf{29.49} &
    \underline{12.76} & \textbf{16.12} & \textbf{13.52} & 9.48 &
    \textbf{7.38} & \underline{18.66} & \textbf{66.52} & \textbf{36.98} &
    \textbf{40.00} & \textbf{26.02} & \textbf{26.20} \\
    \hline
  \end{tabular}}
\end{table*}

\begin{table}[t]
    \centering
    \caption{Occupancy RayIoU on Occ3D-nuScenes validation set.}
    \label{tab:rayiou}
    \scriptsize
    \setlength{\tabcolsep}{5pt}
    \begin{tabular}{l|c|ccc}
        \hline
        Method & mRayIoU & RayIoU@1 & RayIoU@2 & RayIoU@4 \\ \hline
        GaussianOcc\cite{gan2025gaussianocc}
        & 11.85 & 8.69 & 11.90 & 14.95 \\
        GaussianFlowOcc\cite{boeder2025gaussianflowocc}
        & 16.47 & 11.81 & 16.58 & 20.98 \\
        \hline
        LetOccVote-B & \underline{19.86}  & \underline{14.73} & \underline{20.20} & \underline{24.65}  \\
        LetOccVote-L & \textbf{20.9}1 & \textbf{15.80} & \textbf{21.30} & \textbf{25.62} \\
        \hline
    \end{tabular}
\end{table}

\subsection{Ablation Studies}

\subsubsection{Effectiveness of Proposed Components}

Table~\ref{tab:abl} evaluates the effectiveness of Depth Vote and Semantic Vote with IoU and mIoU. Exp. 0 removes both voting mechanisms while keeping the same base pipeline. Adding Depth Vote increases IoU from 50.51 to 51.44 and also improves mIoU, showing that it mainly enhances geometric occupancy prediction while providing additional benefits to semantic prediction. With Semantic Vote, IoU remains nearly unchanged, whereas mIoU improves by 1.02 points. This indicates that Semantic Vote primarily improves semantic discrimination by strengthening reliable semantic supervision and suppressing unreliable pseudo-labels. Combining both modules achieves the best overall performance, with 51.45 IoU and 19.49 mIoU. These results demonstrate that Depth Vote and Semantic Vote provide complementary improvements in geometry and semantics, respectively.

\begin{table}[t]
    \centering
    \caption{Ablation on the proposed Depth Vote and Semantic Vote.}
    \label{tab:abl}
    \begin{tabular}{c|cc|cc}
        \hline
        No.& \makecell{Depth Vote} 
             & \makecell{Semantic Vote} 
             &  IoU&mIoU \\
        \hline
        0 &  &  &  50.51&18.01\\
        1 & \checkmark &  &  51.44&18.69\\
        2 &  & \checkmark &  50.60&19.03\\
        3 & \checkmark & \checkmark &  51.45&19.49\\ \hline
    \end{tabular}
\end{table}


\subsubsection{Ablation on $k$-$W$ Values}

\begin{table}[t]
\centering
\caption{Ablation on $(k,W)$ configurations.}
\label{tab:kw}
\begin{tabular}{c|ccccc}
\toprule
$k$ & 2 & 2 & 3 & 3 & 4 \\
$W$ & 3 & 5 & 5 & 7 & 7 \\
\midrule
IoU  & 49.58 & 51.45 & 51.33 & 51.87 & 51.78 \\
mIoU & 19.14 & 19.49 & 19.24 & 19.61 & 19.43 \\
Memory(GB) & 14.8 & 19.6 & 19.6 & 24.4 & 24.4 \\
\bottomrule
\end{tabular}
\end{table}

Table~\ref{tab:kw} studies the sensitivity of our consensus mechanism to the temporal window size $W$ and the required number of votes $k$. We report occupancy IoU and mIoU together with the memory footprint of the model and input data under each configuration.

We first examine the effect of the temporal window size. With $k=2$, increasing $W$ from 3 to 5 improves IoU from 49.58 to 51.45 and mIoU from 19.14 to 19.49. A similar trend is observed with $k=3$, where increasing $W$ from 5 to 7 further improves IoU from 51.33 to 51.87 and mIoU from 19.24 to 19.61. These results show that a broader temporal window provides richer cross-frame evidence for consensus, although at the cost of increased memory consumption.

For a fixed $W$, a larger $k$ does not necessarily yield better performance. A stricter consensus criterion allows fewer observations to reach agreement, thereby reducing the effective coverage of consensus-based refinement. Accordingly, $(k,W)=(2,5)$ outperforms $(3,5)$, while $(3,7)$ performs better than $(4,7)$.

Although $(k,W)=(3,7)$ achieves the best overall accuracy, its improvement over $(2,5)$ is relatively small while requiring substantially more memory. Considering memory efficiency, we therefore adopt $(k,W)=(2,5)$ as the default Base configuration.

\subsection{Robustness to Pseudo-Label Quality}

Weakly supervised occupancy prediction depends on the quality of pseudo-label supervision. We therefore evaluate the robustness of LetOccVote under different pseudo-label qualities, using Exp. 0 in Table~\ref{tab:abl} as the baseline configuration. For depth supervision, we use Metric3D-S/L/G to generate pseudo-depth maps. For semantic supervision, we consider GDINO SwinT + SAM ViT-B, GDINO SwinB + SAM ViT-H, and SAM3~\cite{carion2026sam3}. Here, GDINO (Grounding DINO)~\cite{liu2024groundingdino} provides open-vocabulary object detections, while SAM converts the detected regions into pixel-level masks. We further assess the resulting pseudo-label quality using sparse LiDAR depth measurements and LiDAR segmentation annotations. Across these models, stronger pseudo-label generators consistently yield lower depth errors or higher semantic mIoU, confirming that they provide supervision of different quality levels. In each experiment, we vary only one pseudo-label source while keeping the other fixed.


Table~\ref{tab:robust_depth} evaluates different depth models with GDINO SwinB + SAM ViT-H fixed for semantic supervision. LetOccVote consistently improves all three settings, with mIoU gains of 1.19--1.55 points and higher occupancy IoU throughout. Interestingly, Metric3D-G achieves higher mIoU but slightly lower IoU than Metric3D-L for both the baseline and LetOccVote. More accurate depth may improve object boundaries and category-wise localization, thereby benefiting semantic prediction. Meanwhile, irregular surface depths of large classes such as vegetation and manmade can lead to fragmented features during depth-based lifting despite their relatively compact voxel annotations, which may limit the corresponding gain in occupancy IoU.

Table~\ref{tab:robust_sem} evaluates different semantic pseudo-labels with Metric3D-L fixed. LetOccVote again improves all three settings. The largest gain occurs with GDINO SwinT + SAM ViT-B, where mIoU improves by 2.27 points, while gains of 1.48 and 1.45 points are retained with GDINO SwinB + SAM ViT-H and SAM3, respectively. These results show that LetOccVote remains effective across different semantic pseudo-label qualities, with larger benefits under noisier supervision.



\begin{table}[t]
    \centering
    \caption{Robustness to depth pseudo-label quality.
    Semantic supervision is fixed to GDINO SwinB + SAM ViT-H.
    AbsRel is measured against sparse LiDAR points.}
    \label{tab:robust_depth}
    \small
    \setlength{\tabcolsep}{3.5pt}
    \begin{tabular}{@{}l|c|cc|cc@{}}
        \hline
        \multirow{2}{*}{Depth Model}
        & \multirow{2}{*}{\makecell{Pseudo\\AbsRel$\downarrow$}}
        & \multicolumn{2}{c|}{Baseline}
        & \multicolumn{2}{c}{LetOccVote} \\
        & & IoU & mIoU & IoU & mIoU \\
        \hline
        Metric3D-S & 0.247 & 44.53 & 15.98 & 44.73 & 17.17 \\
        Metric3D-L & 0.206 & 50.51 & 18.01 & 51.45 & 19.49 \\
        Metric3D-G & 0.155 & 50.33 & 19.09 & 51.31 & 20.64 \\
        \hline
    \end{tabular}
\end{table}

\begin{table}[t]
    \centering
    \caption{Robustness to semantic pseudo-label quality.
    Depth supervision is fixed to Metric3D-L. mIoU is assessed on sparse LiDAR-seg annotations.}
    \label{tab:robust_sem}
    \small
    \setlength{\tabcolsep}{3.5pt}
    \begin{tabular}{@{}l|c|cc|cc@{}}
        \hline
        \multirow{2}{*}{Semantic Model}
        & \multirow{2}{*}{\makecell{Pseudo\\mIoU$\uparrow$}}
        & \multicolumn{2}{c|}{Baseline}
        & \multicolumn{2}{c}{LetOccVote} \\
        & & IoU & mIoU & IoU & mIoU \\
        \hline
        \makecell[l]{GDINO SwinT +\\ SAM ViT-B}
            & 35.16 & 50.23 & 17.12 & 50.91 & 19.39 \\
        \makecell[l]{GDINO SwinB +\\ SAM ViT-H}
            & 38.16 & 50.51 & 18.01 & 51.45 & 19.49 \\
        SAM3
            & 44.56 & 50.56 & 18.45 & 51.54 & 19.90 \\
        \hline
    \end{tabular}
\end{table}



%% file: sec/5_conclusion.tex
\section{Conclusion}
\label{sec:conclusion}

We present LetOccVote, an effective framework for learning 3D occupancy from 2D pseudo-label supervision. We observe that weakly supervised occupancy learning is sensitive to noisy pseudo-labels and therefore exploit consensus across observations to improve supervision reliability. Specifically, Depth Vote improves geometric consistency and refines depth supervision, leading to more effective feature aggregation and geometric representation. Semantic Vote learns the feature patterns of reliable and contested pseudo-labels, enabling different semantic segments to be selectively emphasized or suppressed during training. Extensive experiments demonstrate that LetOccVote achieves improved accuracy and robust performance for 3D occupancy learning under 2D pseudo-label supervision. Nevertheless, challenges remain for ambiguous semantic categories and severely noisy pseudo-labels. Future work may leverage stronger visual or vision-language models to establish more reliable associations between semantic classes and visual features. This could further improve label efficiency for 3D occupancy learning.

\section{Acknowledgments}
\label{sec:acknowledgments}

This work was supported in part by Guangdong Basic and Applied Basic Research Foundation under Grant 2023A1515110729, and Shenzhen Science and Technology Program under Grants 20231128093642002, JCYJ20220818103006012, KJZD20240903100202004, ZDCY20250901100405006, and ZDCY20250901103359008.